# A-SEE2.0: Active-Sensing End-Effector for Robotic Ultrasound Systems with Dense Contact Surface Perception Enabled Probe Orientation Adjustment

Yernar Zhetpissov, Xihan Ma, Kehan Yang, *Member, IEEE*, Haichong K. Zhang, *Member, IEEE*

*Abstract*—**Conventional freehand ultrasound (US) imaging is highly dependent on the skill of the operator, often leading to inconsistent results and increased physical demand on sonographers. Robotic Ultrasound Systems (RUSS) aim to address these limitations by providing standardized and automated imaging solutions, especially in environments with limited access to skilled operators. This paper presents the development of a novel RUSS system that employs dual RGB-D depth cameras to maintain the US probe normal to the skin surface, a critical factor for optimal image quality. Our RUSS integrates RGB-D camera data with robotic control algorithms to maintain orthogonal probe alignment on uneven surfaces without preoperative data. Validation tests using a phantom model demonstrate that the system achieves robust normal positioning accuracy while delivering ultrasound images comparable to those obtained through manual scanning. A-SEE2.0 demonstrates 2.47 ± 1.25 degrees error for flat surface normal-positioning and 12.19 ± 5.81 degrees normal estimation error on mannequin surface. This work highlights the potential of A-SEE2.0 to be used in clinical practice by testing its performance during in-vivo forearm ultrasound examinations.**

*Index Terms* — **Computer vision for medical robotics, medical robots and systems, robotics and automation in life sciences, sensor-based control.**

## I. INTRODUCTION

ULTRASOUND (US) imaging has established itself as a pivotal tool in medical diagnostics, finding applications in obstetrics [1], cardiology [2], guiding interventional procedures [3], and assisting in radiotherapy treatments [4]. Renowned for their cost-effectiveness, real-time capabilities, and safety, US examinations, however, come with physical demands that can be taxing for sonographers [5]. Professionals are required to exert significant pressure when positioning the US probe on the patient's body, often having to fine-tune the probe's position for an optimal image in an ergonomically challenging manner.

Moreover, the outcomes of US examinations are highly operator-dependent, influenced by factors such as scan locations, probe orientations, and contact force at the scan location. Achieving consistent and reliable results demands highly skilled personnel with substantial experience. Unfortunately, such resources are becoming scarce globally [6]. Additionally, the close proximity between sonographers and patients poses infection risks, especially in the context of contagious diseases [7].

To address these challenges, researchers have explored the development of an autonomous Robotic US Systems (RUSS). These innovative systems utilize robot arms to manipulate the US probe, alleviating sonographers of physical burdens. The number of RUSS-related publications has grown exponentially, reaching 125,110 from 2001 to 2022 [8]. Importantly, RUSS allow for remote diagnosis, eliminating the need for direct contact between sonographers and patients [2]. The robot arm can precisely control the probe's pose (position and orientation) and the applied force, ensuring high motion precision and, consequently, securing examination accuracy and repeatability.

Most autonomous RUSS implementations adopt a two-step strategy. First, a scan trajectory is defined based on preoperative data, such as Magnetic Resonance Imaging (MRI) or a vision-based point cloud of the patient's body [9,10]. In the second step, the robot travels along the trajectory, continuously updating the probe pose and applied force based on intraoperative inputs, such as force/torque sensing and real-time US images.

However, challenges remain in the acquisition of diagnostically meaningful US images. Involuntary patient movements, errors in scan trajectory registration to patient, and the highly deformable nature of the skin pose significant hurdles. Real-time adjustment of probe positioning and orientation is crucial, especially for maintaining appropriate acoustic coupling between the transducer and the body. Proper probe orientation ensures a clearer visualization of pathological clues in US images, but achieving near real-time reliable adjustment remains a formidable challenge in the field.

This paragraph of the first footnote will contain the date on which you submitted your paper for review, which is populated by IEEE. This work was supported by Worcester Polytechnic Institute internal funding and the National Institutes of Health under Grant DP5 OD028162 and Grant R01 DK133717. *(Yernar Zhetpissov and Xihan Ma are co-first authors) (Corresponding author: Haichong K. Zhang.)*

This work involved human subjects or animals in its research. Approval of all ethical and experimental procedures and protocols was granted by the institutional research ethics committee at Worcester Polytechnic Institute, under Application No. IRB-21-0613.

Yernar Zhetpissov, Xihan Ma, and Kehan Yang are with the Department of Robotics Engineering, Worcester Polytechnic Institute, Worcester, MA 01609 USA (e-mail: ytzhetpissov@wpi.edu, xma4@wpi.edu).

Haichong K. Zhang is with the Department of Robotics Engineering, Worcester Polytechnic Institute, Worcester, MA 01609 USA, and also with the Department of Biomedical Engineering, Worcester Polytechnic Institute, Worcester, MA 01609 USA (e-mail: hzhang10@wpi.edu).

### A. Related Works

The orthogonal placement of the US probe with respect to the patient body is considered as natural pose for maximum acoustic coupling. Thus, RUSS imaging related works attempt to place the transducer perpendicular to the skin surface [11,12,13,14]. RGB-depth (RGB-D) cameras have been widely utilized for medical procedures [15], and in RUSS particularly [10,12,13,14,16]. Despite their popularity, they are mainly used for offline preoperative scan planning and probe placement.

Most of the previous works on RUSS focused on offline preoperative collection of the patient's region of interest to be scanned for path planning. G. Ma *et al.* [17] used eye-in-hand RGB-D camera for RUSS of a forearm and relied on the point cloud surface of a forearm for orientation adjustment during the scan. In addition, Zhang *et al.* [18] proposed a spatial and temporal probe orientation compensation strategy. However, the calculation of the normal vector relied only on the average of five vectors and sampling time is relatively high (0.2s) for real-time applications which may compromise the system accuracy and responsiveness.

To achieve real-time orthogonal positioning simultaneously in-plane and out-of-plane, X. Ma *et al.* [19] proposed the active-sensing end-effector (A-SEE) for RUSS using four range lasers. Although being a cost-effective and time-efficient approach for self-normal-positioning for lung ultrasound (LUS), usage of fixed laser sensors is limited by relatively flat large areas [20]. The perception capability of such an end-effector is limited due to its invariance to the sensed surface texture and complexity, which limits its ability to scan narrow anatomical regions, such as upper and lower extremities and the neck. To summarize, the usage of end-effector-placed RGB-D depth cameras for probe orientation control is explored only to a limited extent. Additionally, more sophisticated perception methods for the area under US examination may be essential to advance learning-based RUSS in an innovative and impactful manner, offering an additional source of information for skill learning. Thus, a novel real-time, full-field-of-view, depth perception-based approach is needed.

### B. Contributions

The current work builds upon the further improvement of the A-SEE [19]. The means of sensing mechanism that can perceive the tissue texture to be probed is lacking in the previous works. Although A-SEE robot was able to accurately orient US probe relative to the probed tissue during a relatively simple lung sonography task, due to the large probing area and small required probe inclination angles, it would fail in cases like forearm scanning, where even a single laser sensing outside the human tissue area would result in incorrect orientation due to its invariance to the sensed surface. In this work, we aim to lay the foundation for increased robustness and intelligence of the autonomous RUSS systems by addressing the essential problem of reliable alignment of the US probe relative to a patient body in real-time. This is accomplished by introducing a novel robot end-effector with an US probe, equipped with two RGB-D cameras that continuously perceive the body surface near the probe. The required rotation adjustments towards the desired surface-probe orientation are computed based on the fused depth camera point clouds. A human-shared autonomy system is implemented where the probe can automatically land on a patient's body, autonomously adjust its orientation and contact force during imaging, and be slid on the patient's body and rotated about its long axis via teleoperation. The main contributions of this paper are summarized as follows:

- We propose the second generation (the first generation was introduced in [19]) of the compact and cost-effective end tool for RUSS, referred to as the active-sensing end-effector 2.0 (A-SEE2.0). A-SEE was designed to provide real-time information on the rotation adjustment required for achieving normal positioning. A-SEE is the first form factor that allows simultaneous in-plane and out-of-plane real-time US probe orientation adjustment. A-SEE2.0 further implements dense RGB-D sensing to, for the first time, enable full field of view perception of a probed tissue, allowing for reliable optimal orthogonal probe placement in real-time.

- We integrate A-SEE2.0 with a RUSS system and implement a complete US imaging workflow to demonstrate its self-normal-positioning capability.

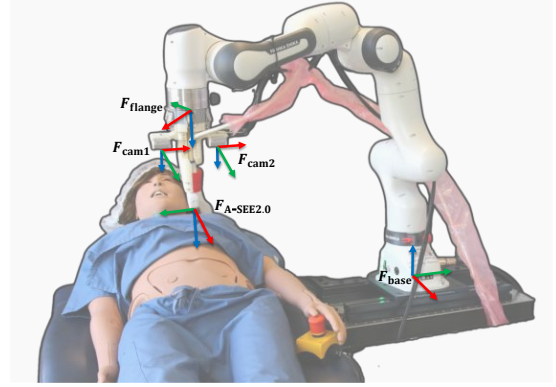

**Fig. 1.** A-SEE2.0 setup with coordinate frames. $F_{base}$ is robot base frame; $F_{flange}$ is frame end-effector attached to; $F_{cam1}$ and $F_{cam2}$ are RGB-D cameras' frames; $F_{A-SEE2.0}$ is US probe tip frame.

## II. MATERIALS AND METHODS

This section describes the implementation details of A-SEE2.0 for RUSS integration. The proposed A-SEE2.0-integrated RUSS (shown in Fig. 1) performs intraoperative probe self-normal-positioning with contact force control. The shared control scheme facilitates teleoperated sliding of the probe along the patient's body surface and enables rotation of the probe about its axis during imaging.

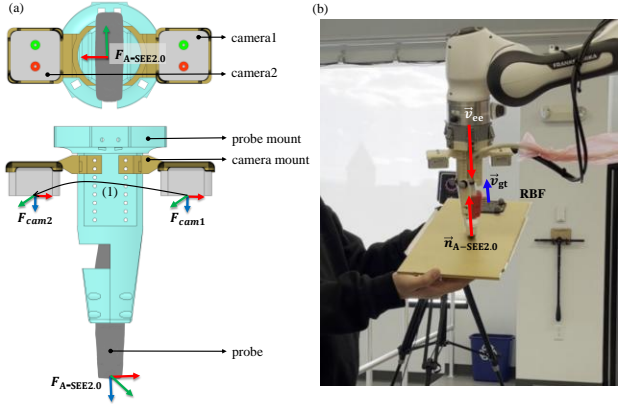

**Fig. 2.** ASEE2.0 design and probe normal positioning vectors definition. (a) Bottom- and side-views of CAD model. Cameras are on the opposite sides of out-of-plane US imaging plane. (1) is the transformation $F_{cam2}^{cam1}$ between $F_{cam1}$ and $F_{cam2}$; (b) Flat plane following experiment and ASEE2.0 vector and frame definitions.

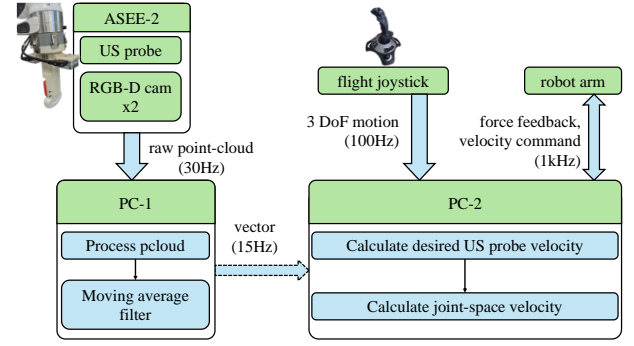

**Fig. 3.** RUSS equipped with ASEE2.0 imaging workflow diagram. Green boxes represent hardware, blue boxes - calculations, arrows - data and command communication.

## A. Technical Approach Overview

The implementation of the system was accomplished through the following tasks: **Task 1)** A-SEE2.0 fabrication, RGB-D cameras calibration and integration with RUSS; **Task 2)** Orientational positioning via the 2-DoF rotational motion of the probe; **Task 3)** US probe contact force control via the 1-DoF translational motion; **Task 4)** Teleoperation control using a joystick for the remaining 3-DoF.

A-SEE2.0's perception component consists of two short-range RGB-D cameras (D405, RealSense, Intel, USA) attached to side-mounted brackets (see Fig. 2). These cameras are chosen due to their sub-millimeter accuracy at close-range and compactness with an ideal depth sensing range between 7 cm and 50 cm. The fused point cloud from the two cameras, located at both sides, provides a full 360-degree view of the probed target area. Point cloud processing and normal estimation are performed on workstation "PC-1" (see Fig. 3).

The wired US probe (C1-6 Curved Array Probe, General Electric, USA) is encapsulated in a 3D-printed mount. A clamping mechanism connects A-SEE2.0 to the 7-DoF robotic manipulator (Panda Research 3, Franka Emika, Germany). The robot motion control and the teleoperation pipeline are implemented on workstation "PC-2". System communication is established using Robot Operating System (ROS).

The robot's self-normal-positioning is activated before landing on the patient body. In-plane and out-of-plane rotation of the US probe, ensuring real-time orthogonal placement, is based on the fused point cloud. Constant force adaptation allows automatic translational motion in probe z-axis. The remaining 3 DoF (probe z-axis rotation and x- and y-axis translations along the body) are controlled by the operator via a 3-DoF flight joystick (PXN-2113-SE, PXN, Japan).

## B. System Calibration

Knowing the target body position in 3D space relative to the world coordinates is crucial for accurate US probe placement tasks. For our setup, intrinsic and extrinsic camera calibrations were necessary. The intrinsic calibration ensures accurate acquisition of the point cloud data relative to each camera. The extrinsic calibration, involving camera-to-camera and eye-in-hand calibration, allows the accurate localization of the point cloud data relative to the robot base frame.

For intrinsic calibration, each camera was calibrated using the on-chip self-calibration feature of Intel RealSense software, utilizing A4-sized printed textured pattern target. Since the relative placement of the two cameras is kept simple, the ground-truth relative location of the two RGB-D cameras, based on the 3D printed mount dimensions, was used for camera-to-camera calibration. This was defined as a static homogeneous transformation $F_{cam2}^{cam1}$, in the system.

The Park and Martin eye-in-hand calibration method [21] was used to obtain a single homogeneous transformation between the left RGB-D camera and the end-effector of the robot. Twenty forward kinematics transformations and corresponding images of a 5x6 checkerboard printed pattern were collected while manually operating the robot end-effector using the image calibration toolbox [22]. It resulted in point cloud positioning relative to the last link of the robot. The forward kinematics of the robot then allows the transformation of a point cloud to the robot base frame. Thus, the normal vector of the probed surface relative to the end-effector is known, allowing for further self-normal positioning.

## C. Probe Normal-Positioning

The probe's angular positioning in A-SEE2.0 relies on the fused point cloud generated from the RGB-D cameras and robot kinematics. This section describes the point cloud processing pipeline, normal vector estimation, and the probe normal positioning by actuating the robot arm to align with the estimated normal vector.

Point cloud processing is required for precise patient skin sensing and real-time normal vector estimation. Fig. 2(b) illustrates schematically the normal vector estimation. The raw point cloud from each camera is cropped along the z-axis of $F_{cam1}$ and $F_{cam2}$, preserving points in [0.02, 0.25] cm range. The proximity to the sensed surface inherent in US procedures leads to an excessively high density of points near the sensed surface. To reduce redundancy, the data is downsampled to 35000 points per camera.

The fused point cloud is obtained by concatenating the point clouds from cam1 and cam2 using the rigid body transformation $F_{cam2}^{cam1}$, calculated in Section II-B.

The Point Cloud Library (PCL) is used for further denoising of point clouds and normal estimation. The points corresponding to the US probe were box-cropped using the known probe and mount dimensions, as well as their relative position to the cameras. Then, voxel grid downsampling, followed by statistical outlier removal were performed.

For point cloud data streamed at every timestamp, the normal vector is computed as the average of local normals over a 10 x 10 cm rectangular region centered under the probe tip $F_{A-SEE2.0}$. Due to the ambiguity of the estimated vector direction, the z-axis component of $\vec{n}_{A-SEE2.0}$ is kept positive as follows:

$$\vec{n}_{A-SEE2.0} = [x, y, z] = \begin{cases} [-x', -y', -z'] & z' < 0 \\ [x', y', z'] & z' \geq 0 \end{cases}, \quad (1)$$

where $\vec{n}_{A-SEE2.0}$ is the vector normal to the probed surface, $[x', y', z']$ is the original output vector of normal estimator.

A moving average filter is applied to the estimated normal vector components independently to ensure smooth motion. The size of the moving window was empirically chosen equal to 7 as a trade-off between robustness and minimal latency.

With accurate normal estimation in real-time, A-SEE2.0 can be integrated with the robot to enable motion that tilts the US probe, aligning it with the normal direction of the skin surface. As depicted in Fig. 2(b), upon normal positioning of the probe, the angular difference between the vectors $\vec{n}_{A-SEE2.0}$ and $F_{A-SEE2.0}$ frame's $z$-direction vector $\vec{v}_{ee}$ is minimized. Thus, we address **Task 2** by simultaneously applying in-plane rotation $\omega_y$, and out-of-plane rotation $\omega_x$. The angular velocities about the two axes at timestamp $t$ are given by a PD control law

$$\begin{bmatrix} \omega_x \\ \omega_y \end{bmatrix} = \begin{bmatrix} K_p & K_d & 0 & 0 \\ 0 & 0 & K_p & K_d \end{bmatrix} \begin{bmatrix} proj_{xz}\vec{v}_{ee} \\ \frac{\Delta proj_{xz}\vec{v}_{ee}}{\Delta t} \\ proj_{yz}\vec{v}_{ee} \\ \frac{\Delta proj_{yz}\vec{v}_{ee}}{\Delta t} \end{bmatrix}, \quad (2)$$

where $K_p$ and $K_d$ are empirically tuned control gains; $proj_{xz}\vec{v}_{ee}$ and $proj_{yz}\vec{v}_{ee}$ are projections of $\vec{v}_{ee}$ on $xz$ and $yz$ planes, respectively;
$\Delta proj_{xz}\vec{v}_{ee} = proj_{xz}\vec{v}_{ee}(t) - proj_{xz}\vec{v}_{ee}(t-1)$;
$\Delta proj_{yz}\vec{v}_{ee} = proj_{yz}\vec{v}_{ee}(t) - proj_{yz}\vec{v}_{ee}(t-1)$;
$\Delta t$ is control time interval. The angular velocity control response can reach 30 Hz.

### D. US Probe Contact Force Control

To prevent loose contact between the probe and the skin, which can cause acoustic shadows in the image, a force control strategy is essential to maintain consistent pressure throughout the imaging process. This strategy also ensures that the probe maintains gentle contact with the body, prioritizing patient safety. We adapted a force control method from A-SEE that regulates the linear velocity along the z-axis within the A-SEE2.0 framework. Velocity adaptation follows a two-stage approach, handling landing and scanning motions separately. During landing, the probe's velocity asymptotically decreases as it approaches the body surface. During scanning, the velocity is adjusted based on deviations between the measured desired force values. Thus, the velocity along the z-axis at time $t$ is calculated as

$$v_{fz}(t) = w \cdot v + (1-w) \cdot v_{fz}(t-1), \quad (3)$$

where $w$ is a constant in the range (0,1) to maintain the smooth velocity profile, and $v$ is computed as

$$v = \begin{cases} K_{p1}(\tilde{d} - \min(\vec{d_z})) & \min(\vec{d_z}) \geq \tilde{d} \\ K_{p2}(\tilde{F} - \bar{F}_z) & \min(\vec{d_z}) < \tilde{d} \end{cases}, \quad (4)$$

where $\vec{d_z}$ is the vector of the z-components of the processed point cloud; $\bar{F}_z$ is the measured tip force along the z-axis of $F_{A-SEE2.0}$, estimated from joint torque readings provided by the robot software; $\tilde{F}$ is the desired reference contact force; $K_{p1}$, $K_{p2}$ are empirically determined gains; $\tilde{d}$ is the threshold distance distinguishing the landing stage from the scanning stage, set to the distance from camera lens to the tip of the probe (150 mm).

### E. RUSS equipped with A-SEE2.0 Imaging Workflow

The A-SEE2.0 system, providing 7-DoF control of the US probe, integrates self-normal positioning, contact force regulation, and teleoperation capabilities. Fig. 3 illustrates the overall A-SEE2.0 system block diagram for RUSS.

During the landing stage, the probe is gradually lowered until it reaches the reference contact force. Once contact is established, the operator gains control over the probe's lateral movements in the $x$-$y$ plane and its rotation along the long axis. Using joystick teleoperation, the operator can slide the probe along the patient's body surface, generating end-effector velocities in the $F_{A-SEE2.0}$ body frame. These velocities are then transformed into joint velocities as follows:

$$\dot{q} = J(\vec{q})^\dagger \begin{bmatrix} \vec{v}^{base} \\ \vec{\omega}^{base} \end{bmatrix}, \quad (5)$$

where $J(\vec{q})^\dagger$ is the Moore-Penrose pseudoinverse of the robot Jacobian.

The body velocities are transformed into linear and angular velocities in the space base frame as:

$$\begin{bmatrix} \vec{v}^{base} \\ \vec{\omega}^{base} \end{bmatrix} = \begin{bmatrix} R_{A-SEE2.0}^{base} & [p]R_{A-SEE2.0}^{base} \\ 0_{3\times3} & R_{A-SEE2.0}^{base} \end{bmatrix} \begin{bmatrix} \vec{v}^{A-SEE2.0} \\ \vec{\omega}^{A-SEE2.0} \end{bmatrix}. \quad (6)$$

The commanded probe velocity is expressed as:

$$\begin{bmatrix} \vec{v}^{A-SEE2.0} \\ \vec{\omega}^{A-SEE2.0} \end{bmatrix} = \begin{bmatrix} v_{tx} & v_{ty} & v_{tz} & \omega_{nx} & \omega_{ny} & \omega_{nz} \end{bmatrix}^T. \quad (7)$$

Throughout the procedure, US images are continuously streamed to the workstation. After the operator completes the scanning, the robot automatically returns to its original home configuration.

## III. EXPERIMENT SETUP

### A. Validation of Self-Normal-Positioning

Three experiments were designed to evaluate the self-normal positioning and surface reconstruction capabilities of A-SEE2.0. Experiment A1 (expt. A1) evaluates the system's ability to track the normal vector on a flat surface (ideal case) and its response time. Experiment A2 (expt. A2) assesses normal vector estimation on complex uneven surfaces, simulating anthropomorphic structures. Experiment A3 (expt. A3) evaluates system calibration and point cloud fusion accuracy by measuring phantom surface reconstruction performance.

In expt. A1, reflective optical markers were tracked in real-time using a motion capture system (Vantage, Vicon Motion Systems Ltd, U.K.). Four trackers were placed on the robot end-effector, coplanar with its $xy$ plane, to determine $\vec{v}_{ee}$, while another four trackers were placed on the flat surface to define $\vec{v}_{gt}$ (See Fig. 2(b)). The error is defined as the angular difference between $\vec{v}_{ee}$ and $\vec{v}_{gt}$. In the experiment, rotations about the x-, y-axis, and z-axis translation of $F_{A-SEE2.0}$ were enabled. Four optical markers on the flat surface formed a rigid body frame (RBF in Fig. 2(b)) with its z-axis perpendicular to the surface, representing $\vec{v}_{gt}$. The plane board was held against the US probe and rotated in four phases. During each phase, the operator altered $\vec{v}_{gt}$ by tilting the plane to a certain angle. Meanwhile, the robot was aligning $\vec{v}_{ee}$ and $\vec{v}_{gt}$ using the self-normal-positioning feature. Once the robot reached a steady state, the next phase was introduced. The four phases were repeated four times, during which synchronized $\vec{v}_{ee}$ and $\vec{v}_{gt}$ values were recorded. The normal positioning errors were calculated.

In expt. A2, phantom upper torso (COVID-19 Live Lung Ultrasound Simulator, CAE Healthcare™, USA) mimicking patient body texture was utilized as the experimental subject for evaluation of local surface normals estimation. The experiment was set up to examine A-SEE2.0's ability to estimate accurate surface normals. Target regions were chosen to match anterior, anterior-lateral, and lateral regions of the chest examined during a lung US procedure as per BLUE lung US protocol [23]. This allows testing of normal angle estimation across a variety of tilt angles. The phantom was scanned using the 3D Scanner app to obtain the ground truth body surface representation. To register the scanned phantom point cloud with A-SEE2.0 point cloud, four fiducials were placed on the phantom surface. The target regions were manually selected on the left side of the torso, as shown in Fig. 4(d). Targets 1, 4, 7, 10 are in the anterior region; targets 3, 6, 9, 12 are in the lateral region; targets 2, 5, 8, 11 are in the middle region. The targets were chosen from the scanned point cloud. Each target region represents a circular area with a radius of 3 cm and ground truth normal $\vec{v}_{gt}$ are estimated as local normals to each circular area. The robot was manually moved to six positions (see Fig. 4(e)) with RGB-D cameras facing the phantom to collect A-

SEE2.0 point clouds, ensuring visibility of the whole region of interest with targets and the fiducials. Source normal vectors were calculated as described in Section II-C for each region of interest. The point clouds were registered using fiducial correspondences between the scanned and A-SEE2.0 point clouds. The obtained registration frames were used to transform target points to $F_{A-SEE2.0}$ frame for each capturing position. The angle errors of the normal estimation were calculated accordingly for each target. The mean angle error across 6 robot positions was recorded.

In expt. A3, a breast elastography phantom (Part Number 1552-01, CIRS) was used to evaluate point clouds fusion by estimating surface reconstruction accuracy. The Chamfer Distance (CD) metric was chosen as a commonly used surface similarity measure in computer vision research [25]. CD effectively captures the overall geometric similarity of two point clouds and is insensitive to slight local misalignment, making it suitable for this evaluation. The robot end-effector was placed above the phantom and was capturing the whole phantom surface. Similar to expt. A2, breast phantom ground truth surface representation was acquired using 3D Scanner app and registered with A-SEE2.0 fused point cloud using four fiducials. The ground truth surface representation of the breast phantom was acquired and registered with the A-SEE2.0 fused point cloud in the same manner as in expt. A2.

### B. Validation of US Image Quality

To demonstrate the superiority of the A-SEE2.0 system over the previous generation A-SEE, a human subject study was conducted to evaluate its ability to acquire high-quality images in forearm US scanning, a scenario involving a small surface area. The study was approved by the institutional research ethics committee at the Worcester Polytechnic Institute (No. IRB-21-0613). Written informed consent was given to the volunteers prior to all test sessions. Three volunteers were recruited for the right forearm US procedure. The overall setup is shown in Fig. 5(a).

The operator teleoperated the probe to the center of the target's forearm, while A-SEE2.0 dynamically optimized its orientation. US images were continuously acquired during the whole procedure. The probe was teleoperated to scan the entire length of the subject's forearm. Ten representative images in which the subjects' artery walls were visible were selected. Since forearm arteries are commonly examined in forearm US [24], we used their visibility, quantified by the contrast-to-noise ratio (CNR), as a measure of image quality. For a given a rectangular region of interest (ROI) on the landmark, CNR is defined as:

$$CNR = \frac{|\mu_{roi} - \mu_{bg}|}{\sqrt{\sigma_{roi}^2 + \sigma_{bg}^2}}, \tag{8}$$

where $\mu_{roi}$ and $\sigma_{roi}$ are the mean and the standard deviation of the pixel intensities in the ROI, $\mu_{bg}$ and $\sigma_{bg}$ are the mean and the standard deviation of the pixel intensities in the image background. The CNR of the vessel wall was calculated for US images acquired using A-SEE2.0. The ROI

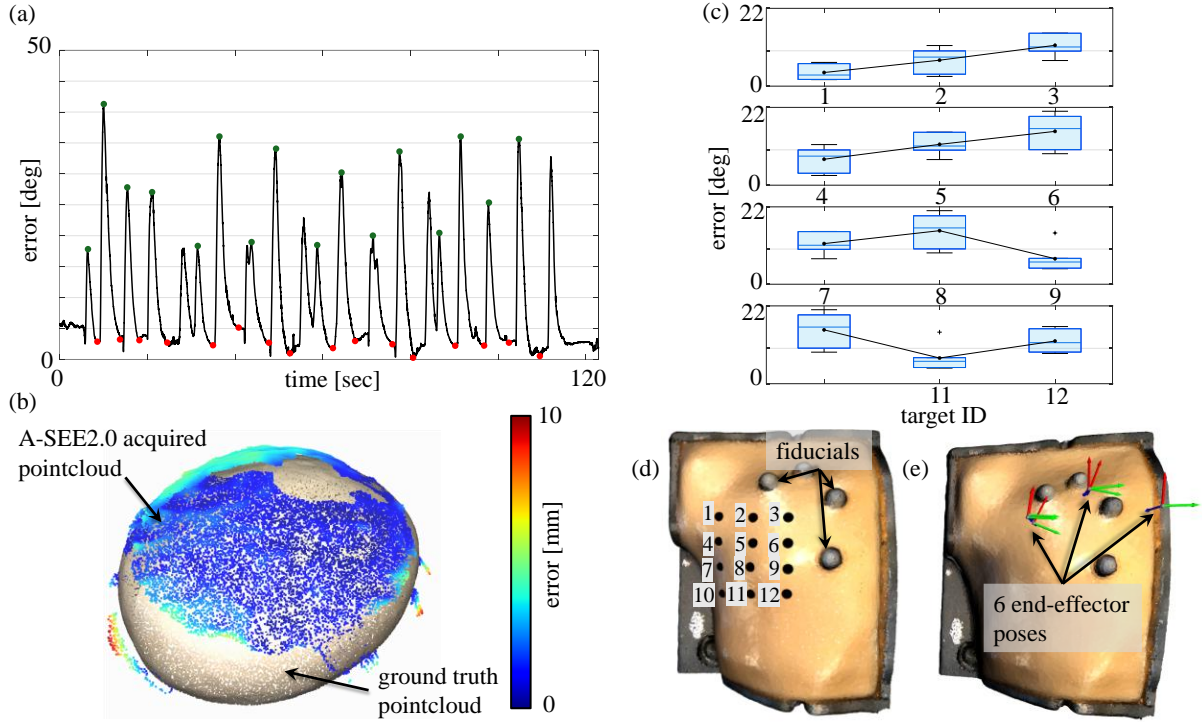

**Fig. 4.** Self-normal-positioning validation experiment. (a) Recorded normal positioning errors on the flat surface over time. The red dots indicate the residual errors (i.e., the minimum normal positioning error achievable) after normal positioning; the green dots indicate maximum recorded errors for each phase; (b) Breast phantom surface perception error heatmap of the fused point cloud for expt. A3. Biege colored points represent scanned ground-truth; (c) The normal estimation errors calculated at 12 targets on lung phantom during expt. A2; (d) Setup for expt. A2 depicting 12 targets and fiducials configuration; (e) $F_{A-SEE2.0}$ frame location snapshots during expt. A2 data collection process.

area was fixed at 10 x 10 pixels for CNR computation.

### C. Contact Force Control Validation

In the aforementioned human study, we validated the effectiveness of the force control strategy. The force exerted at the probe tip, recorded by the robot's sensors, was sampled at 30 Hz throughout the experiment. The target contact force was empirically set to a constant 3.5 N. To quantify control performance, we used the force control error, defined as the difference between the measured and desired forces.

## IV. RESULTS

### A. Validation of Self-Normal-Positioning

Fig. 4(a) shows the results of flat surface tracking in expt. A1 using A-SEE2.0. The error peaked when the operator rotated the plane in four distinct directions, then decreased rapidly as the robot aligned the probe with the normal. The mean normal positioning error across sixteen manually positioned planes was $2.47 \pm 1.25$ degrees, calculated by averaging the residual errors (see red dots in Fig. 4(a)). Response time was computed as a measure of the robot's responsiveness to the sudden changes in the imaging surface orientation. It is defined as the time interval between the previous peak and the corresponding valley. The average response time measured was $3.34 \pm 0.56$ seconds.

Compared to the same experiment on A-SEE ($4.17 \pm 2.24$ degrees and $3.67 \pm 0.84$ seconds), A-SEE2.0 improved both average normal positioning error and response time by 40.77% and 8.99%, respectively. The experiment demonstrated that A-SEE2.0 improved performance in the ideal case of tracking a flat surface. The results suggest that the system can adapt to sudden geometric changes in the probed surface.

Fig. 4(c) shows the normal estimation errors when sensing the phantom in expt. A2. Errors when scanning the anterior targets (1, 4, 7, 10), anterior-lateral targets (2, 5, 8, 11), and lateral targets (3, 6, 9, 12) were $13.46 \pm 10.72$ degrees, $8.43 \pm 1.43$ degrees, and $12.10 \pm 0.80$ degrees, respectively. The mean error for all 12 targets was $12.19 \pm 5.81$ degrees, indicating generally acceptable self-normal-positioning performance across the upper torso regions. A-SEE2.0 exhibited higher accuracy for anterior-lateral targets, as most positions were aligned perpendicularly to this region. A-SEE

## TABLE I
### IMAGE QUALITY FOR HUMAN SUBJECTS FOREARM US

| Subject | Age | BMI | Average Pain level | CNR |
|---|---|---|---|---|
| 1 | 30 | 22.4 | 1.67 | $3.29 \pm 0.69$ |
| 2 | 32 | 23.6 | 0 | $4.02 \pm 0.80$ |
| 3 | 28 | 29.2 | 0 | $4.41 \pm 1.39$ |

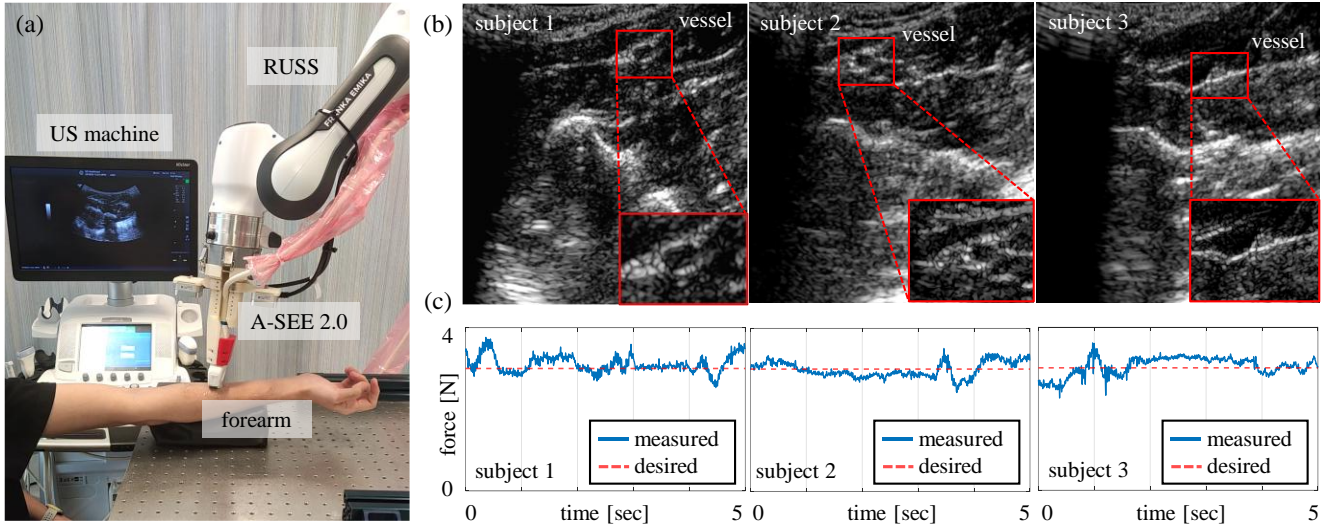

**Fig. 5.** Human subjects experiment. (a) Human subject experimental setup. (b) Example US images of forearms. Red boxes show chosen vessel walls as ROI for CNR metric of image quality evaluation step. (c) Force profiles during forearm US experiment for all 3 subjects.

produced results for a similar experiment, with the key difference that the robot performed self-normal-positioning to the mannequin surface rather than normal vector estimation. The approximate normal estimation for the mannequin is improved by 16.9%. The experiment was redesigned due to the lack of a ground-truth phantom surface. Self-normal positioning and force control were disabled in this experiment to isolate the evaluation of normal direction estimation capability.

Fig. 4(b) presents the heatmap of the reconstructed surface. The average CD was calculated as 7.5. As expected, the error was highest at outer points, which represent very steep surfaces relative to the camera's point of view. Overall, point cloud fusion produced a reliable surface reconstruction outcome, validating the system's calibration accuracy.

### B. Validation of US Image Quality

The assessment of US image quality and contact force control was carried out during human subject forearm US experiment. Fig. 5(b) illustrates exemplary US images captured during the experiment with ROI being depicted. Vessel walls have been chosen as ROI, and CNR of the regions have been calculated. Table I shows the image quality results corresponding to each subject. Mean CNR for 3 recruited subjects is $3.91 \pm 0.57$. Pain level (scale of 10) associated with the probe pressure during the experiment has been queried 3 times for each subject throughout the imaging. 2 out of 3 subjects reported 0 level of pain, 1 subject reported a minor discomfort of 1.67, showing the system's safety performance. Overall, vessel walls were clearly visible and differentiable from the surrounding tissue.

### C. Validation of Contact Force Control

Fig. 5(c) presents the probe's z-direction force profiles recorded during imaging for each subject. The force errors remained mostly within $\pm 0.5$ N, with an average force error of $0.041 \pm 0.215$ N. The results demonstrate that the contact force controller effectively maintained the applied force near the target value during real forearm US imaging.

## V. DISCUSSION AND CONCLUSIONS

A novel end-effector A-SEE2.0 for RUSS with real-time probe normal-positioning capability has been presented. This work extends the previous version of the active-sensing end-effector by incorporating RGB-D cameras for a more comprehensive perception of the scanned surface. This feature could be advantageous for RUSS by enabling dense sensing of the contact area, complementing existing RUSS approaches. The use of in-hand RGB-D cameras may enhance the understanding of probe-tissue interactions during US scanning, potentially enabling a more intelligent and generalizable framework for autonomous RUSS. Although only orthogonal positioning has been evaluated in the paper, omnidirectional orientation control of the US probe is theoretically possible using A-SEE2.0. This flexibility is valuable for meeting the requirements of various imaging applications (e.g., liver, heart, gallbladder) [26, 27]. Given the limitations of RUSS with the previous A-SEE iteration, which was constrained by a fixed sensing region of a 35 mm radius, the addition of depth data expands its scanning capabilities beyond simple front-chest LUS to more advanced applications, such as forearm US. A-SEE2.0 demonstrated improvements in normal positioning and estimation, with a 40.77% reduction in error for the ideal case ($2.47 \pm 1.25$ degrees for a flat surface) and a 16.9% improvement in the more realistic case ($12.19 \pm 5.81$ degrees for a mannequin). Additionally, A-SEE2.0 does not suffer from optical occlusion and surface deformation as eye-to-hand based RUSS works [10, 13]. Consistently high CNR results across subjects with varying BMI values demonstrate the generalizability of the A-SEE2.0 pipeline for populations with diverse body habitus [28].

This study is limited to evaluating only the normal-positioning capability and image quality during forearm US experiments on human subjects. Although only normal-positioning experiments have been conducted, the system shows strong potential for real-time optimal US probe placement and, potentially, learning-based RUSS approaches [29]. Thus, imitation learning capabilities will be integrated into A-SEE2.0 to evaluate its performance in specific scenarios, such as gallbladder, liver, and heart US, bringing RUSS closer to clinical implementation.